\documentclass[aps,pre,groupedaddress,twocolumn]{revtex4}

\usepackage{amsthm}
\usepackage{amsmath}
\usepackage{amssymb}
\usepackage{graphicx}
\usepackage{color}
\usepackage{bm}
\usepackage{subfigure}
\usepackage{multirow}
\usepackage{latexsym}
\usepackage{array}
\newtheorem{theorem}{Theorem}
\newtheorem{lemma}[theorem]{Lemma}

\begin{document}
\title{Generalization of k-means Related Algorithms}
\author{Yiwei Li \\ \href{mailto: mrhutumeng@gmail.com}{mrhutumeng@gmail.com}}

\begin{abstract}
   This article briefly introduced Arthur and Vassilvitshii's work on \textbf{k-means++} algorithm and further generalized 
   the center initialization process. It is found that choosing the most distant sample point from the nearest center as new center can mostly have the same effect as the 
   center initialization process in the \textbf{k-means++} algorithm.
\end{abstract}

\maketitle

\section{Background and Introduction}
\label{secIntro}
Clustering is one of the classic unsupervised machine learning problems. It has been demonstrated to be NP-hard even with only two clusters \cite{Drineas2004}. 
In 1982, Lloyd \cite{Lloyd1982} gave a local search solution to solve this problem, which is one of ``the most popular clustering algorithms used in scientific 
and industrial applications'' \cite{Berkhin2002}, which is also known as \textbf{k-means}. The total error $\phi$ is monotonically decreasing, and the process will always terminate since the number of 
possible clusterings is finite ($\leqslant k^n$, where $n$ is the total number of sample points) \cite{Arthur2006}. However, the accuracy of the \textbf{k-means} 
algorithm cannot be always good enough. In fact, many examples show that the algorithm generates arbitrarily bad clusterings ($\frac{\phi}{\phi_{\rm{OPT}}}$ is proved 
to be unbounded even if $n$ and $k$ are fixed, where $\phi_{\rm{OPT}}$ is the optimal total error) \cite{Arthur2006}. Furthermore, the final clusterings strongly 
depend on the initial setup of the cluster centers. 
\textbf{k-means++} algorithm propose a way to choose random starting centers with very specific probabilities \cite{Arthur2006}, which guarantees the upper-bound of 
the total error expectation by $E[\phi]\leqslant8(\rm{ln}$~$k+2)\phi_{\rm{OPT}}$ for any set of data points \cite{Arthur2006} without sacrifice the fast 
computation speed and algorithm simplicity. In particular, ``\textbf{k-means++} is never worse than $O(\rm{log}$~$k)$-competitive, and on very well formed data sets, 
it improves to being $O(1)$-competitive'' \cite{Arthur2006}.

This article is organized as follows. In section \ref{k-means}, the traditional \textbf{k-means} and the \textbf{k-means++} algorithms are introduced based on reference~\cite{Arthur2006}.
In section \ref{alternatives}, the relation between \textbf{k-means} and \textbf{k-means++} is illustrated, and generalize the initialization 
process of the \textbf{k-means++} algorithm which indicates that to select most distant sample point from nearest center as new center can have the same (or very similar) effect as 
randomly select new center from the entire weighted sample space.

\section{Existing Algorithms}
\label{k-means}

Suppose we are given an integer $k$ and a set of $n$ data points $\mathcal{X}\subset\mathbb{R}^d$ \cite{Arthur2006}. The goal is to select $k$ centers $\mathcal{C}$ 
so as to minimize the potential function (total error)
\begin{equation}
\phi = \sum\limits_{x\in\mathcal{X}}\min\limits_{c\in\mathcal{C}}\lVert x-c \rVert^2 \,.\nonumber
\end{equation}

In this report, I will use the same notation as in Ref~\cite{Arthur2006}: $\mathcal{C}_{\rm{OPT}}$ represents the optimal clustering and $\phi(\mathcal{A})$ represents 
the contribution of $\mathcal{A}\subset\mathcal{X}$ to the potential 
\begin{equation}
\phi(\mathcal{A}) = \sum\limits_{x\in\mathcal{A}}\min\limits_{c\in\mathcal{C}}\lVert x-c \rVert^2 \,.\nonumber
\end{equation}
In general, \textbf{k-means} algorithm has four steps \cite{Arthur2006}: \\
\phantom{x}\hspace{2ex}1.~Randomly choose $k$ initial centers $\mathcal{C}=\{c_1,\dots,c_k\}$. \\
\phantom{x}\hspace{2ex}2.~For each $i\in\{1,\dots,k\}$, set the cluster $\mathcal{C}$ to be the set of points in $\mathcal{X}$ that are closer to $c_i$ than they are to $c_j$ for all $j\neq i$. \\
\phantom{x}\hspace{2ex}3.~For each $i\in\{1,\dots,k\}$, set $c_i$ to be the center of mass of all points in $\mathcal{C}_i:c_i=\frac{1}{|\mathcal{C}_i|}\sum\limits_{x\in\mathcal{C}_i}x$. \\
\phantom{x}\hspace{2ex}4.~Repeat step 2 and 3 until $\mathcal{C}$ no longer changes.

\begin{lemma}
Let $S$ be a set of points with center of mass $c(S)$, and let $z$ be an arbitrary point. Then \\ 
\phantom{x}\hspace{2ex}$\sum\limits_{x\in S}\lVert x-z \rVert^2 - \sum\limits_{x\in S}\lVert x-c(S) \rVert^2 = |S|\cdot\lVert c(S)-z \rVert$.
\label{lemma1}
\end{lemma}

The Lemma~\ref{lemma1} quantifies the contribution of a center $c$ to the cost improvement in a \textbf{k-means} step as a function of the distance it moves \cite{Peled2005}. 
Specifically, if in a \textbf{k-means} step a $k$-clustering $\mathcal{S}=(S_1,\dots,S_k)$ is changed to the other $k$-clustering $\mathcal{S}^{\prime} = (S_1^{\prime},\dots,S_k^{\prime)}$, 
then the total change of potential function

\begin{align} \label{loss_change}
 \phi(\mathcal{S}) - \phi(\mathcal{S}^{\prime}) \geqslant \sum\limits_{j=1}^k |S^{\prime}_j|\cdot\lVert c(S^{\prime}_j) - c(S_j) \rVert^2 \,.
\end{align}
The reason that loss function has a no-less-than sign rather other an equal sign is Lemma~\ref{lemma1} only consider the improvement resulting from step 3 of 
\textbf{k-means} algorithm in which the centers are moved to the centroids of their clusters \cite{Peled2005}. However, there is an additional gain from 
reassigning the points from step 2 of \textbf{k-means} algorithm \cite{Peled2005}. Therefore, \textbf{k-means} algorithm guarantees the potential function monotonically 
decreases over each iteration before reaching the optimal clusterings when initial centers are given. 

Let $D(t)$ denote the shortest distance from a data point $x$ to the closest center we have already chosen. Then, the \textbf{k-mean++} algorithm is \cite{Arthur2006}: \\
\phantom{x}\hspace{2ex}1a. Choose an initial center $c_1$ uniformly at random from $\mathcal{X}$. \\
\phantom{x}\hspace{2ex}1b. Choose the next center $c_i$, selecting $c_i=x^{\prime}\in\mathcal{X}$ with probability $\frac{D(x^{\prime})^2}{\sum\limits_{x\in\mathcal{X}}D(x)^2}$. \\
\phantom{x}\hspace{2ex}1c. Repeat step 1b until we have chosen a total of $k$ centers. \\
\phantom{x}\hspace{2ex}2-4. Proceed as with the standard \textbf{k-means} algorithm. \\
The weighting used in step 1b is called ``$D^2$ weighting''.
The Ref~\cite{Arthur2006} proved an important result as follows:
\begin{theorem}
If $\mathcal{C}$ is constructed with \textbf{k-means++}, then the corresponding potential function $\phi$ satisfies $E[\phi]\leqslant8(\rm{ln}$~$k+2)\phi_{\rm{OPT}}$.
\label{theorem1}
\end{theorem}

\section{Alternative Approaches and Their Relations}
\label{alternatives}
The \textbf{k-means++} algorithm demonstrates that during the center initialization process, it is much better to select centers with probability proportional to their square distance with nearest 
existing center. It is equivalent to say that the \textbf{k-means++} is the weighted initialized \textbf{k-means}. In a more general case, we can 
tuning the portion of sample points which can be randomly selected as a new center. In particular, a hyper-parameter $\alpha \in (\epsilon, 1]$ is set to determine the most distant $\alpha \times N$ ($N$ is the size 
of sample points and $\epsilon = \frac{1}{N}$) points from their nearest existing centers, and then select the new center from them instead of the entire dataset. 
The two extreme cases are (1) when $\alpha=\epsilon$ so that we deterministically choose the most distant point from its nearest center (it saves computation time during center initialization with sacrifice of 
not considering the distribution of the dataset), and (2) when $\alpha=1$ so that we go back to exact \textbf{k-means++} algorithm. 
The $\alpha$ values $\{\epsilon, 0.5, 1.0\}$ are tested on different datasets (e.g. wines and Spam datasets in \cite{UCI}) for different $k$ numbers (e.g. 3, 10, 20) compared with traditional \textbf{k-means} algorithm. 
No matter the computation time, average potential or minimal potential are very similar or exactly same (the computation time is similar as traditional \textbf{k-means} but 
average and minimal potential is one magnitude lower), even exclude the randomness of first initial center (see Table~\ref{wines_k10},~\ref{spam_k10} and~\ref{spam_k20}). This might indicates that the main advantage of the \textbf{k-means++} algorithm can be explained or replaced 
by selecting the most distant point from the nearest center.

Except for testing the potentials, it is also possible to evaluate the accuracy of the clustering for some specific dataset. For instance, the Iris dataset in \cite{UCI} has three classes. When using \textbf{k-means}-related algorithms, 
it will mainly give two distinct clustering ways: one is same as the ground truth classification; the other is group Virginia and Versicle into one cluster and split Samoset into two clusters. Table~\ref{iris_k3} shows the ratio of 
obtaining the correct classification for different algorithms.
\begin{table}[hbt!]
\centering
\begin{tabular}{ | c | c | c | c | }
\hline
 Algorithm & Avg Potential & Min Potential & Time \\ 
 \hline
 \textbf{k-means} & 3.81$\times$10$^5$ & 2.18$\times$10$^5$ & 1\\  
 \textbf{k-means++} & 2.53$\times$10$^5$ & 2.18$\times$10$^5$ & 1.05\\
 $\alpha=\epsilon$  & 2.55$\times$10$^5$ & 2.18$\times$10$^5$ & 1.06\\
 No random & 2.54$\times$10$^5$ & 2.18$\times$10$^5$ & 1.09\\
 $\alpha=0.5$ & 2.54$\times$10$^5$ & 2.18$\times$10$^5$ & 1.10 \\
\hline
\end{tabular}
\caption{Results for wines dataset \cite{UCI}, $k=10$, $n=5000$. ``No random'' represents that the first initial center is selected by the most distant sample point from a sample point uniformly randomly selected 
from the entire sample space and the following initial centers are selected as the most distant sample points from the nearest existing centers. }
\label{wines_k10}
\end{table}
\vfill
\begin{table}[hbt!]
\centering
\begin{tabular}{ | c | c | c | c | }
\hline
 Algorithm & Avg Potential & Min Potential & Time \\ 
 \hline
 \textbf{k-means} & 4.19$\times$10$^8$ & 1.75$\times$10$^8$ & 1\\  
 \textbf{k-means++} & 9.35$\times$10$^7$ & 7.70$\times$10$^7$ & 1.05\\
 $\alpha=\epsilon$  & 9.35$\times$10$^7$ & 7.70$\times$10$^7$ & 1.06\\
 No random & 9.62$\times$10$^7$ & 7.70$\times$10$^7$ & 1.05\\
 $\alpha=0.5$ & 9.23$\times$10$^7$ & 7.70$\times$10$^7$ & 1.09\\
\hline
\end{tabular}
\caption{Results for Spam dataset \cite{UCI}, $k=10$, $n=1200$.}
\label{spam_k10}
\end{table}
\vfill
\begin{table}[hbt!]
\centering
\begin{tabular}{ | c | c | c | c | }
\hline
 Algorithm & Avg Potential & Min Potential & Time \\ 
 \hline
 \textbf{k-means} & 2.58$\times$10$^8$ & 1.50$\times$10$^8$ & 1\\  
 \textbf{k-means++} & 2.50$\times$10$^7$ & 2.14$\times$10$^7$ & 1.35\\
 $\alpha=\epsilon$  & 2.50$\times$10$^7$ & 2.14$\times$10$^7$ & 1.35\\
 No random & 2.46$\times$10$^7$ & 2.14$\times$10$^7$ & 1.42\\
 $\alpha=0.5$ & 2.46$\times$10$^7$ & 2.14$\times$10$^7$ & 1.43\\
\hline
\end{tabular}
\caption{Results for Spam dataset \cite{UCI}, $k=20$, $n=1200$.}
\label{spam_k20}
\end{table}
\vfill
\begin{table}[hbt!]
\centering
\begin{tabular}{ | c | c | c | }
\hline
 Algorithm & Accuracy & Time \\ 
 \hline
 \textbf{k-means} &  0.08 & 1\\  
 \textbf{k-means++} &  0.91 & 1.00\\
 $\alpha=\epsilon$  &  0.91 & 0.99\\
 No random & 0.91 & 1.02\\
 $\alpha=0.5$ &  0.91 & 1.01\\
\hline
\end{tabular}
\caption{Results for Iris dataset \cite{UCI}, $k=3$, $n=10000$.}
\label{iris_k3}
\end{table}

\section{Summary}
In this article, the existing \textbf{k-means} and \textbf{k-means++} algorithms are briefly introduced. In center initialization process, the former only considers the samples density distribution 
while the latter also take the distance into account to modify the sample density distribution. Afterwards, the initialization process is generalized and couple of alternative approaches are compared. 
It is found that choosing the most distant sample point from the nearest existing center can mostly have the same effect as considering the entire sample space.

\end{document}